\documentclass{article}

\usepackage[final]{corl_2019} % Uncomment for the camera-ready ``final'' 
\usepackage{times}
\usepackage{epsfig}
\usepackage{graphicx}
\usepackage{amsmath}
\usepackage{amssymb}
\usepackage{subcaption}
\usepackage{multirow}

\title{Multi-Frame GAN: Image Enhancement \\for Stereo Visual Odometry in Low 
Light}

\author{
Eunah Jung$^\text{1,2*}$ \quad Nan Yang$^\text{1,2*}$ \quad Daniel 
Cremers$^\text{1,2}$\\
$^\text{1}$Technical University of Munich \quad $^\text{2}$Artisense\\
}

\begin{document}
\maketitle

%===============================================================================

% !TeX root = ../main.tex

\begin{abstract}
We propose the concept of a multi-frame GAN (MFGAN) and demonstrate its 
potential as an image sequence enhancement for stereo visual odometry in low 
light conditions. We base our method on an 
invertible adversarial network 
to transfer the beneficial features of brightly illuminated scenes to the 
sequence in poor illumination without costly paired datasets. In 
order to preserve the coherent geometric cues for the translated sequence, we 
present a novel network architecture as well as a novel loss term 
combining temporal and stereo consistencies based on 
optical flow estimation. We demonstrate that the enhanced sequences improve the 
performance of state-of-the-art feature-based and direct stereo 
visual odometry methods on both synthetic and real datasets in challenging 
illumination. We also show that MFGAN outperforms other state-of-the-art image 
enhancement and style transfer methods by a large margin in terms of visual 
odometry.
\end{abstract}


\begin{figure}[h]
	\centering
	\includegraphics[width=.8\linewidth]{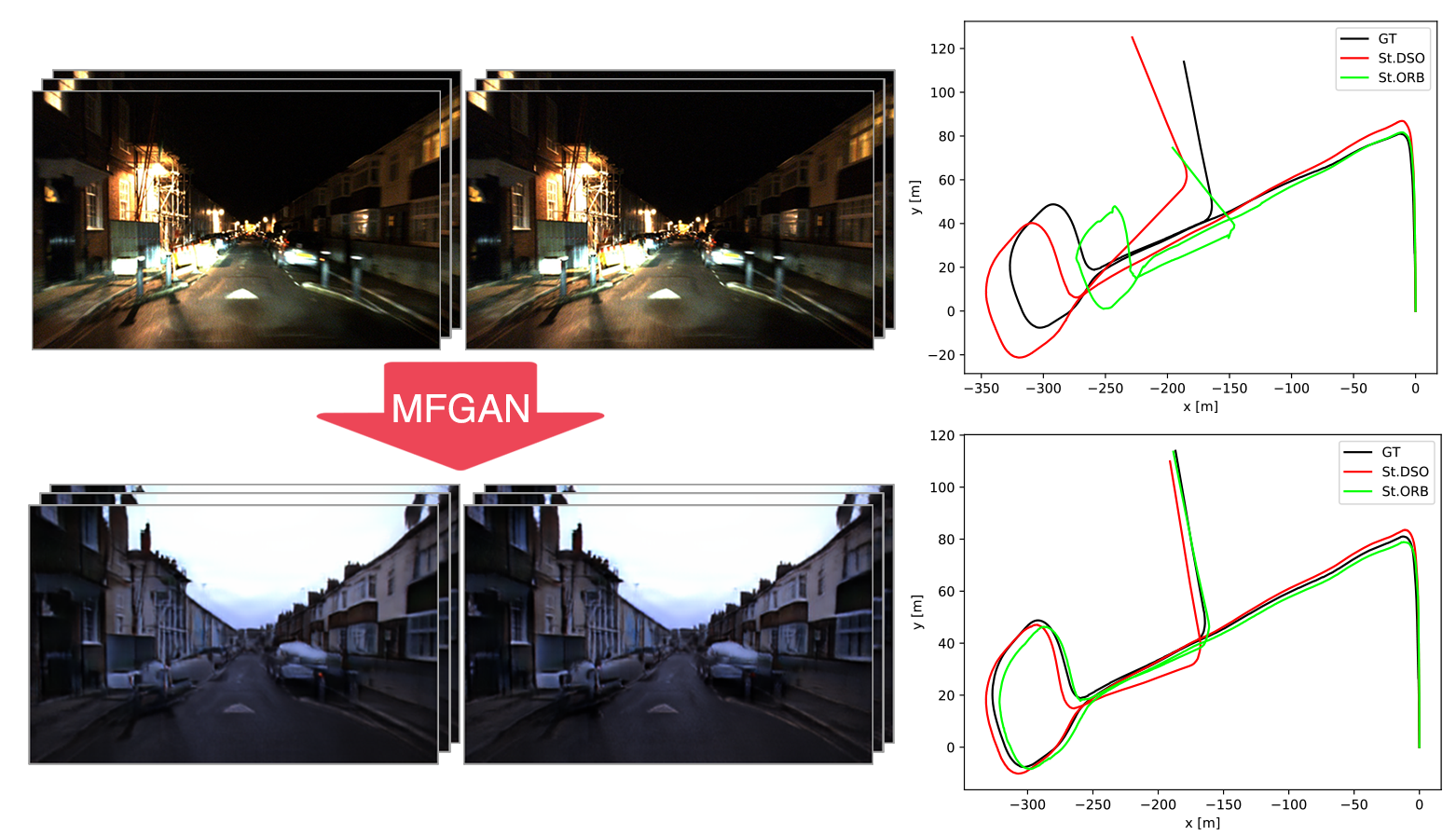}
	\caption{We propose Multi-Frame GAN (MFGAN) for 
		stereo VO in challenging low light environment. 
		The MFGAN takes two consecutive stereo image pairs and outputs the 
		enhanced stereo images while preserving temporal and stereo 
		consistency. On 
		the right side, the estimated trajectories 
		by the state-of-the-art stereo feature-based VO method Stereo 
		ORB-SLAM
		and the state-of-the-art direct VO method Stereo DSO are presented. 
		Due to the low image gradient, dynamic lighting and halo, Stereo DSO 
		and Stereo ORB-SLAM cannot achieve good tracking accuracy in the 
		night scene. With the translated images from MFGAN, the performance of 
		both methods is notably improved.}
\end{figure}

\let\thefootnote\relax\footnotetext{*These two authors contributed equally. 
Correspondence to: \texttt{\{jungeu,yangn\}@in.tum.de}}
% !TeX root = ../main.tex

\section{Introduction}

Visual odometry (VO) and simultaneous localization and mapping (SLAM) have been 
actively studied due to their wide usage in robotics, AR/VR and autonomous 
driving. Particularly, stereo VO\cite{ 
mur2017orb,wang2017stereo} delivers more reliable and accurate results 
than monocular systems\cite{engel2018direct, engel2014lsd, mur2017orb} 
by eliminating the scale 
ambiguity\cite{strasdat2010scale,yang2018deep,yang2018challenges}.

\begin{figure}[t]
	\centering
	\includegraphics[width=0.9\textwidth]{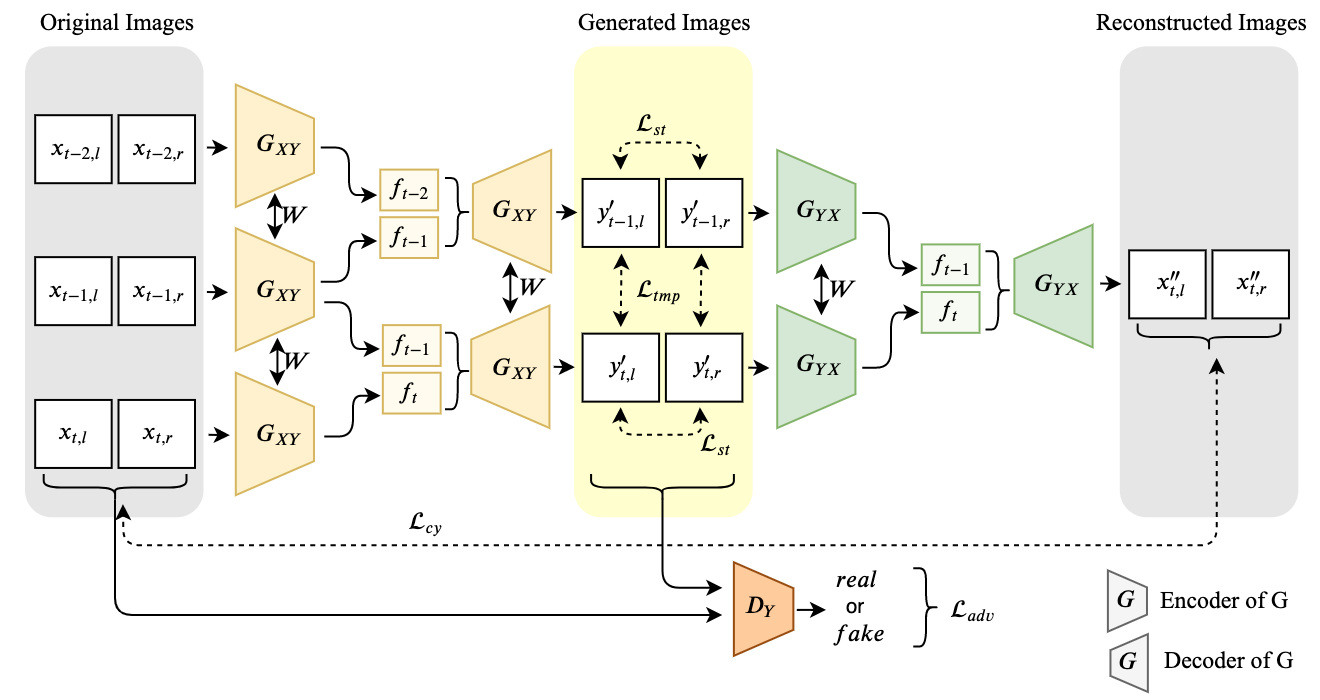}
	\caption{Overview of MFGAN. Only the forward cycle with $G_{XY}$ and $D_Y$ 
	is 
	shown 
		for the simplicity.	The encoder of the generator takes a stereo image 
		pair at each timestamp and the 
		decoder part takes the concatenated feature maps from previous and 
		current
		frames to output the current enhanced images. During 
		training phase, the adversarial loss $\mathcal{L}_{adv}$ is computed by 
		using 
		discriminator network and the cycle consistency loss $\mathcal{L}_{cy}$ 
		compares 
		the original and reconstructed images. The network generates two 
		consecutive fake frames and computes the temporal consistency loss 
		$\mathcal{L}_{tmp}$ 
		between different times and the stereo consistency loss 
		$\mathcal{L}_{st}$ 
		between stereo image pairs.
		%		The weights of the generators are shared 
		%		as above and optimized all together after calculating the total 
		%loss 
		%for both forward and backward 
		%		cycles.
	}
	\label{fig:overview}
\end{figure}

However, tracking the camera pose in poorly-lit conditions, such as night 
driving scenes, which is crucial for autonomous driving, is still a 
challenge 
for current stereo systems. Both, feature-based 
methods~\cite{klein2007parallel,mur2017orb} and direct 
methods~\cite{engel2014lsd,engel2018direct} rely on image gradient-based key 
point extraction, which provides fewer high-quality points in low-light scenes.
Moreover, with dynamic lighting and halo that are abundant in dark scenes, it 
is difficult to track the same key points as the
illumination is changing. To overcome these limitations, several approaches 
have recently been presented, e.g., camera exposure time 
control~\cite{shim2014auto, zhang2017active}, camera model optimization 
~\cite{bergmann2017online,zhang2017active}, and robust feature 
descriptors~\cite{alismail2017direct,pascoe2017nid}. Yet, these approaches 
either need paired datasets for training or address only one of many aspects 
for the challenging night scene.

In this paper, we propose a learning-based sequence enhancement 
method for stereo VO methods, named Multi-Frame GAN (MFGAN). MFGAN makes use of 
Generative Adversarial Networks 
(GANs) to perform a domain transfer from 
bad illumination to good illumination. In this way, manual engineering for 
different aspects causing VO failure at night is avoided.
%By 
%doing so, we can avoid possible engineering steps for every different camera 
%sensors which might lead to extra cost and expect the generalization effect of 
%training the networks on the large dataset.
Based on CycleGAN~\cite{zhu2017unpaired}, we make use 
of \textit{unpaired} data for training, thus avoiding substantial costs for 
pairing or labeling in real-world applications. Yet, CycleGAN transforms images 
independently whereas for stereo VO we need to preserve spatial (i.e., 
inter-camera) and temporal consistency of 
the domain transfer. To this end, we carefully 
design the temporal and stereo consistency loss terms leveraging optical flow 
in order to ensure consistency of the transformed brightness across cameras and 
in time.

%In addition, our method shows the augmented module to estimate pixel-wise 
%uncertainty maps, which explains how reliable the generated appearance is to 
%be 
%under the context of the target domain. This enables us to differentiate 
%reliable and uncertain information for robust performance in real-world 
%applications.
We validate our approach with state-of-the-art VO methods on the synthetic 
indoor New Tsukuba 
dataset~\cite{martull2012realistic} and the challenging outdoor Oxford RobotCar 
datatset~\cite{maddern20171} which contains various illumination situations. 
Specifically, we use a direct method, Stereo 
DSO~\cite{wang2017stereo}, and a feature-based method, Stereo 
ORB-SLAM~\cite{mur2017orb}. To the best of our knowledge, this is the first work 
exploring the potential of GAN-based image translation for VO. And the 
experiments show that our method leads to significant 
improvement in accuracy and robustness for both direct as well as indirect 
methods. 
We also compare MFGAN with other photo 
enhancement~\cite{zuiderveld1994contrast,guo2016lime,deephotoenhancer} and 
image/video translation~\cite{anoosheh2018night,DRIT,li2018learning} methods. 
The results show the superiority of MFGAN for improving VO in the challenging 
lighting condition.

% !TeX root = ../main.tex

\section{Related Work}

\textbf{Robust Visual Odometry (VO).} Feature-based 
methods~\cite{mur2017orb,klein2007parallel} 
rely on feature matching and estimate the camera poses by minimizing the 
re-projection error. Direct methods~\cite{engel2014lsd,engel2018direct} do not 
rely on feature descriptors and directly optimize the photometric error. To 
improve the performance of VO in challenging lighting 
conditions, Pascoe et al.~\cite{pascoe2017nid} proposed a 
direct monocular SLAM algorithm using a newly designed metric considering 
entropies in the frame instead of intensities. Alismail et al. 
\cite{alismail2017direct} introduced a binary feature descriptor for direct VO 
methods for poor light environment. While the classical vision methods have 
been 
actively researched, there are few learning-based methods. Gomez et al. 
\cite{gomez2018learning} trained a neural network with LSTM units using 
synthetic paired datasets to produce enhanced images and evaluated their method 
on real-world static scene. Compared to their method, 
we make use of unpaired datasets and explicitly address temporal and 
spatial coherence using optical flow. We validate our method on 
challenging synthetic as well as real-world datasets.

\textbf{Image and Video Translation.} Gatys et al. 
\cite{gatys2016image} used the pretrained networks to capture the content and 
style respectively and optimize texture transfer. Image translation using 
Generative Adversarial Networks (GANs) \cite{isola2017image, 
zhu2017unpaired, wang2018high} has become very popular due its superior 
performance. Isola \cite{isola2017image} proposed a conditional 
adversarial networks with paired samples in different styles and this work is 
extended to CycleGAN by Zhu et al. \cite{zhu2017unpaired} who suggested two 
pairs of unconditionally trained adversarial networks. One of the applications 
of image translation is retrieval-based visual 
localization~\cite{anoosheh2018night,porav2018adversarial}. They used image 
translation to close the domain gap for matching images from different 
conditions. TodayGAN~\cite{anoosheh2019night} adapts the architecture of 
CycleGAN and improves the performance of image retrieval. Beyond a 
single image 
style transfer, video synthesis into other styles becomes 
an active topic \cite{chen2017coherent, wang2018video, 
gao2018reconet}. Especially, synthesizing video requires temporal consistency 
over the contiguous frames, since estimating individual frames leads to 
flickering 
effect. Therefore, many works use an approach comparing the 
target image and the warped image based on the optical flow to push consistency 
\cite{wang2018video, chen2017coherent}. While we also 
utilize image-warping based on flow field, we design our 
consistency loss function taking into account both stereo-spatial as well as 
temporal context.
% !TeX root = ../main.tex

\section{Multi-Frame GAN for Space-time Consistent Domain Transfer}
Our method, Multi-Frame GAN (MFGAN), is based on the cycle-consistent network 
architecture proposed by Zhu et al.~\cite{zhu2017unpaired}. We extend it into a 
multi-frame 
scheme, such that MFGAN translates a given sequence of stereo images in one 
domain into another domain of sequence. With the proposed temporal and stereo 
consistency terms, MFGAN is able to generate the translated sequence preserving 
the coherence of the input sequence. We implement this coherence with 
differentiable image 
warping using optical flow.
%Therefore, 
%our method aims to generate a current stereo pair by taking two \mbox{-} 
%previous and current \mbox{-} stereo pairs and complete the coherent generated 
%sequence as if it is a real video. 
%Our method, MFGAN, is based on cycle-consistent 
%network architecture 
%\cite{zhu2017unpaired} and translates a given sequence of stereo image pairs $X=\{(X_{i}^{l}, 
%X_{i}^{r})_{i=1}^T\}$ for $T$ times into another domain of 
%sequence $Y=\{(Y_{i}^{l}, Y_{i}^{r})_{i=1}^T\}$ for $T$ times while keeping 
%consistent appearance. We preserve consistency over the entire consistent 
%sequence inspired by the observation that every two consecutive frames in the 
%sequence are coherent. Therefore, our method learns to generate the frame pair 
%$(Y_t^l,Y_t^r)$ by taking two frame pairs $(X_{t-1}^l,X_{t-1}^r)$, 
%$(X_t^l,X_t^r)$ and complete the whole smoothly coherent sequence from frame to 
%frame.
%\subsection{Formulation}

Inspired by~\cite{zhu2017unpaired}, MFGAN consists of two 
generator-discriminator-sets, $\{G_{XY},D_Y\}$ and $\{G_{YX},D_X\}$ where $X$ 
and 
$Y$ denote different image domains, e.g., $X$ for the poor lighting and $Y$ for 
good lighting condition.
The generator $G_{XY}$ translates the input images from the source domain $X$ 
to the target domain $Y$, and the discriminator $D_Y$ aims to distinguish 
between the original images from the domain $Y$ and the translated fake images.
Likewise, the other set, $\{G_{YX},D_X\}$, functions the same 
but with opposite domains. We denote the part with $G_{XY}$ and $D_X$ as the 
forward cycle and the part involving $G_{YX}$ and $D_Y$ as the backward cycle.

%$G_{XY}$ translates the input images from the source 
%domain $X$ to the target domain of $Y$. $G_{YX}$ translates back 
%the generated images from the domain of $Y$ to $X$. The discriminators $D_Y$ 
%and $D_X$ aim to distinguish between the original image from certain domain 
%and the translated image.

The overview of our networks in the training phase is shown in 
Figure \ref{fig:overview}. For simplicity, we present the method 
regarding the forward cycle and skip the backward cycle. 
In the training phase, the generator $G_{XY}$ takes \textit{three} pairs of 
stereo images and generates \textit{two} pairs of 
stereo images for the target domain,
\begin{equation}
y'_{t-1} = G_{XY}(x_{t-2},x_{t-1}) \quad \text{and} \quad y'_{t} = 
G_{XY}(x_{t-1},x_{t}).
\end{equation}
where $x_{t-2}, x_{t-1}, x_t \in X$ are the original image pairs and $y'_{t-1}, 
y'_t \in Y$ 
are the fake image pairs. Note that a stereo pair $x_t$
means $(x_{t,l}, x_{t,r})$ for the left and right images. The 
generator $G_{YX}$ takes $(y'_{t-1}, y'_{t})$ as the input 
and reconstructs the images back in the source domain $X$,
\begin{equation}
x''_t = G_{YX}(y'_{t-1}, y'_{t}).
\label{eq:recons_img}
\end{equation}
where  $x''_t \in X$ is the reconstructed stereo image.
%For the simplicity, we denote $G_{XY}(X_{t-2},X_{t-1})$, 
%$G_{XY}(X_{t-1},X_{t})$, $G_{YX}(Y_{t-1}, Y_{t})$ as $G_{XY}^{t-2, t-1}$, 
%$G_{XY}^{t-1, t}$,  
%$G_{YX}^{t-1, t}$, respectively.

%The encoders and the decoders share the weights as shown in 
%Figure \ref{fig:overview}, and the translated image $y'_{t-1}$ and $ y'_t$
%share the input image $x_{t-1}$. This characteristic structure naturally 
%lets the networks learn to utilize the previous frame to have the 
%current frame consistent with it.

%\subsubsection{Adversarial Loss}
\textbf{Adversarial Loss.} The adversarial loss 
$\mathcal{L}_{adv}$~\cite{goodfellow2014generative, zhu2017unpaired, porav2018adversarial, 
anoosheh2018night}
 is defined as:
\begin{equation}
\begin{gathered}
\mathcal{L}_{adv} = \mathcal{L}_{gen} + \mathcal{L}_{disc},\\
\mathcal{L}_{gen} = (D_Y(y'_t) - 1)^2,\\
\mathcal{L}_{disc} = (D_Y(y_t)-1)^2 + (D_Y(y'_t))^2
\end{gathered}
\end{equation}
\label{eq:loss_adv}
The discriminator is trained to distinguish the given real and fake inputs 
correctly 
through $\mathcal{L}_{disc}$, while the generator aims to synthesize 
as realistic image in domain $Y$ as possible by minimizing $\mathcal{L}_{gen}$.

%\subsubsection{Image Consistency Loss}
\textbf{Image Consistency Loss.} The image consistency loss term computes how 
similar two images are. We use a linear combination of L1 loss and 
single scale SSIM~\cite{wang2004image} as the measurement:
\begin{equation}
\mathcal F(a,b) = 
\alpha\frac{1-\text{SSIM}(a,b)}{2} + (1-\alpha)| 
a-b|,
\end{equation}
where $\alpha$ is set to 0.8. Then, the cycle consistency loss is formed using 
the above image similarity metric,
\begin{equation}
\mathcal{L}_{cy}=\mathcal F(x_t, x''_t)
\label{eq:cycle_loss}
\end{equation}
where $x''_{t}$ is from Equation \ref{eq:recons_img}. This loss term resolves 
unconstrained difficulties due to unpaired datasets by reconstructing back the 
generated images into the original image domain. 

Additionally, we introduce the temporal consistency loss $\mathcal{L}_{tmp}$ 
and the stereo consistency loss $\mathcal{L}_{st}$ to jointly optimize the 
image coherence over the multiple temporal and stereo-spatial frames. For 
temporally 
neighboring frames, we warp the images $y'_{t-1}$ into $y'_{t}$ using the 
estimated 
optical flow $W_{t-1}^{t}$, and for stereo frames warp the 
right image $y'_{t,r}$ into $y'_{t,l}$ similarly making use of the estimated optical 
flow $W_r^l$. The temporal consistency loss $\mathcal{L}_{tmp}$ and the stereo 
consistency loss $\mathcal{L}_{st}$ are then formed as below,
\begin{equation}
\mathcal{L}_{tmp}=\mathcal F(\omega_{t-1}^t(y'_{t-1}), y'_t)
\label{eq:tmp_loss}
\end{equation}
\begin{equation}
\mathcal{L}_{st}=\mathcal F(\omega_{r}^l(y'_{t,r}), y'_{t,l})
\label{eq:st_loss}
\end{equation}
where $\omega_{t-1}^t (\cdot)$ is the backward warping function using 
the optical flow $W_{t-1}^t$, and $\omega_{t-1}^t (x_{t-1})$ gives the warped 
images 
of $x_{t-1}$ into $x_t$.

Finally, the total loss $\mathcal{L}$ integrates the adversarial loss, cycle 
consistency loss, temporal consistency as well as stereo consistency loss, 
\begin{equation}
\mathcal{L}=\lambda_{adv}\mathcal{L}_{adv}+\lambda_{cy}\mathcal{L}_{cy}+
\lambda_{tmp}\mathcal{L}_{tmp}+
  \lambda_{st}\mathcal{L}_{st}
\label{eqn:totalloss}
\end{equation}
where $\lambda$ is the weight for each corresponding loss term. In summary, 
this loss function extends the cycle loss of~\cite{zhu2017unpaired} by a 
temporal and a stereo consistency loss thereby assuring that the resulting 
domain transfer preserves a spatio-temporal regularity. Note that 
we compute $\mathcal{L}_{adv}$, $\mathcal{L}_{cy}$ and $\mathcal{L}_{tmp}$ for 
the left and right image in a stereo pair correspondingly and each loss term 
includes both translation direction between generators, e.g. from the 
domain $X$ into the domain $Y$ and vice versa.
%We minimize the total loss 
%$\mathcal{L}$ and get two optimal pairs of a generator and a 
%discriminator. 
%\subsection{Network Architecture}

The generator networks 
contain down-sampling and up-sampling convolutional layers in an 
encoder-decoder scheme. The feature maps of the temporal consecutive two 
neighbor frames are channel-wise concatenated and then fed into the decoder. 
For the discriminators, we adopt the PatchGAN architecture 
\cite{zhu2017unpaired,isola2017image} and exploit its usage through the loss 
terms. Please refer to the 
\href{https://vision.in.tum.de/_media/spezial/bib/jung2019corl-supp.pdf}{supplemental
 material} for the detailed architecture. 

%\todo{
%During the forward pass for $y'_{t-1}$ and $y'_t$ in order to compute temporal 
%consistency, in Figure \ref{fig:overview}, the feature space from $x_{t-1}$ is 
%used 
%for both forwarding and all the weight parameters of networks are shared. 
%Through this characteristic structure, the networks naturally learn to utilize 
%the features of the 
%previous frame to generate the current frame consistent with the previous 
%frame.}

% !TeX root = ../main.tex

\section{Experiments}

We evaluate MFGAN on the synthetic indoor New Tsukuba 
dataset \cite{martull2012realistic} as well as the real outdoor Oxford 
RobotCar dataset \cite{maddern20171}.  
%For evaluation, we use two public datasets, synthetic indoor New Tsukuba 
%dataset \cite{martull2012realistic} and real outdoor Oxford RobotCar 
%dataset \cite{maddern20171} which have various illumination environments along 
%the same trajectory. 
To evaluate the frame consistency, we propose a new metric using the 
optical flow between frames and discuss the consistency 
quantitatively and qualitatively. For the evaluation of VO, we validate the 
performance of two state-of-the-art feature-based and direct stereo VO methods, 
namely Stereo ORB-SLAM \cite{mur2017orb} and Stereo DSO \cite{wang2017stereo}, 
respectively. We run the two VO methods on both original sequences with 
bad lighting conditions (Flashlight for New Tsukuba dataset and 
Night for Oxford RobotCar dataset) and the corresponding translated 
good lighting conditions (\textit{Fluorescent} for New Tsukuba dataset and 
\textit{Day} for Oxford RobotCar dataset) with MFGAN. Both, frame 
consistency as well as VO performance are significantly improved by MFGAN on 
both datasets. In this section, we introduce the extensive evaluation results 
on Oxford RobotCar dataset. Please refer to our 
\href{https://vision.in.tum.de/_media/spezial/bib/jung2019corl-supp.pdf}{supplemental
	material} for the 
evaluation on New Tsukuba dataset.
%\subsection{Datasets}

%\textbf{New Tsukuba} dataset \cite{martull2012realistic} captures static 
%office scene. The dataset contains 1800 stereo image pairs with ground-truth 
%camera pose, disparity maps, occlusion maps and discontinuity maps. The stereo 
%camera travels the fixed trajectory under different lighting such as Daylight, 
%Fluorescent, Lamps and Flashlight. The length of whole trajectory is around 
%50m 
%and the camera poses demonstrate strong rotation change. MFGAN is tested on 
%one pair of unpaired sets, Fluorescent and Flashlight, and used 1000 images as 
%a training set and 800 images for the evaluation. 

\textbf{Oxford RobotCar} dataset \cite{maddern20171} provides a massive amount 
of data collected while driving an approximately 10km route over 1 year in 
different time slots. The dataset recorded almost 20 million images from 6 
cameras mounted on the car, with LIDAR, GPS, INS ground truth, and includes the 
data in 
different weather conditions, seasons and daytimes, e.g. summer, winter, 
rain, night, of 
the same trajectory. We select Day and Night scene where 
Day is the overcast dataset 2015/02/10 and Night is the night-tagged dataset 
2014/12/16 to train MFGAN and evaluate the VO performance. Specifically, we 
choose 10 sub-sequences around 700m long from the 
entire route, such that each sub-sequence includes several characteristics like 
multiple corners, straight-shaped route. Please refer to our 
\href{https://vision.in.tum.de/_media/spezial/bib/jung2019corl-supp.pdf}{supplemental
	material} for the locations of each sub-sequences. We use the 
Seq. 00, 02, 03, 05, 08 
as the training set and Seq. 01, 04, 06, 07, 09 as the testing set. Note that 
there is no overlap segments between training sequences and the testing 
sequences.
Approximately 9000 frames are used for training MFGAN. The train and test 
set are split to be geographically equally distributed.
%\subsection{Implementation Details}

In order to compute the consistency loss functions based on the warped images, 
we use the predicted optical flow by the state-of-the-art flow estimation 
network FlowNet2 \cite{ilg2017flownet}. We measure temporal optical flow and 
stereo spatial optical flow to compute the total consistency loss function. To 
be specific, we use the optical flows predicted in one domain, for example, 
domain $X$, to warp the generated images in another domain e.g. $Y$, and check 
the consistency and vice versa for the other direction. We implement MFGAN with 
PyTorch and train with batch size $1$, $15$ 
epochs and Adam optimizer. The learning rate remains 
$0.0002$ for the first $10$ epochs and decays linearly to zero over 
the next 5 epochs. The weights for each loss term in 
Equation \ref{eqn:totalloss} are set as $\lambda_{adv}$ to 1.0, $\lambda_{cy}$ 
to 10.0, $\lambda_{tmp}$ to 3.0 and $\lambda_{st}$ to 3.0. With the image 
resolution 320x192, MFGAN shows 111 FPS inference performance with NVIDIA 
GeForce GTX 1080.
%For the New Tsukuba datasets, 
%since the dataset is relatively small with 1000 training samples, we train our 
%models with the loss function combining cycle and temporal consistency terms 
%without stereo consistency and validate the effectiveness of additional 
%temporal consistency. For the Oxford RobotCar dataset,

In the following, we show the evaluation results on the testing set. For the 
simplicity, we use $cy$ for the model 
trained with cycle consistency loss, $cy,tmp$ for the model $cy$ with 
additional temporal consistency loss, $cy,st$ for the model $cy$ with 
additional stereo consistency loss, and MFGAN means the proposed model trained 
with cycle, temporal and stereo consistency loss.

\subsection{Frame Consistency}
In this section, we evaluate the visual consistency over multiple 
contiguous frames. We investigate this in terms of both temporal and stereo 
consistency.

\textbf{Quantitative results}. We introduce a metric using optical flow between 
frames as the optical flow is measured by matching the corresponding points of 
two frames by their appearance. Under the assumption that the optical flow 
between two frames has no flaw, the optical flow $W_{t}^{t+2}$ from $t$-th 
frame to $(t+2)$-th frame equals the addition of the optical flow $W_{t}^{t+1}$ 
and $W_{t+1}^{t+2}$ if three contiguous frames at $t$, $t+1$, $t+2$ timestamps 
are temporally consistent in image appearance.

\begin{table}
%	\parbox{.35\linewidth}{
%	\centering
%	\scriptsize
%	\begin{tabular}{c|c|c}
%		\hline
%		& $cy$ &  $cy,tmp$ \\
%		\hline
%		median  & 0.75 & 0.54 
%		\\
%		mean  & 2.20 & 1.68
%		\\
%		\hline
%	\end{tabular}		
%		\caption{EPE $E_{tmp}$ of the New Tsukuba dataset for temporal 
%consistency.}
%		\label{tab:epe_temporal_tsukuba}
%	}
%	\hfill
%	\parbox{.55\linewidth}{
		\centering
		\scriptsize
		\begin{tabular}{c|cc|cc|cc}
			\hline
			& \multicolumn{2}{c|}{$cy$} 
			& \multicolumn{2}{c|}{$cy,tmp$} & 
			\multicolumn{2}{c}{MFGAN} \\ 
			& $E_{tmp}$ & $E_{st}$ & $E_{tmp}$ & $E_{st}$ & $E_{tmp}$ & $E_{st}$ 
			\\
			\hline
			01   & 1.64 & 2.19 & 1.03 & 1.21 & 1.14 & 0.73
			\\
			04   & 2.05& 2.68 & 1.23  & 1.38 & 1.3  & 0.76
			\\
			06   & 1.37 & 1.81 & 0.95  & 1.15 & 1.01 & 0.74
			\\
			07   & 1.76& 2.14 & 1.18 & 1.29 & 1.26 & 0.82
			\\
			09 & 1.92 & 2.46 & 1.16  & 1.35 & 1.2 & 0.83
			\\
			\hline
		\end{tabular}
		\caption{EPE $E_{tmp}$ and $E_{st}$ of the Oxford RobotCar dataset for frame consistency.}
		\label{tab:epe_tmp_st}
		\vspace{-7mm}
%	}
\end{table}

\begin{table}[h]
	\centering
	\scriptsize
	\begin{tabular}{cc|cc||cc|cc|cc|cc}
		\hline
		& & \multicolumn{2}{c||}{Night} &
		\multicolumn{2}{c|}{$cy$} &
		\multicolumn{2}{c|}{$cy,tmp$} &
		\multicolumn{2}{c|}{$cy,st$} &
		\multicolumn{2}{c}{MFGAN}\\ 
		Seq. & & $t_{rel}$ & $r_{rel}$ & $t_{rel}$ & 
		$r_{rel}$& $t_{rel}$ & $r_{rel}$& $t_{rel}$ & $r_{rel}$& $t_{rel}$ & 
		$r_{rel}$\\
		\hline
		\multirow{2}{*}{01}  
		& DSO & 7.16 & 2.91 & X & X & X & X & 10.00 & 2.26 & \textbf{5.41} & 
		\textbf{2.18}\\
		& ORB & X & 4.80 & 21.14 & 2.97 & 78.94 & 26.10 & 81.73 & 26.00 & 
		\textbf{11.06} & \textbf{2.75}\\
		\hline
		\multirow{2}{*}{04}
		& DSO & 24.78 & 5.28 & X & X & X & X & 9.76 & \textbf{2.89} & 
		\textbf{4.75} & 3.20\\
		& ORB & X & 11.00 & 82.73 & 26.22 & 74.70 & 30.74 & 73.03 & 33.43 & 
		\textbf{4.59} & \textbf{3.89}\\
		\hline
		\multirow{2}{*}{06}
		& DSO & 9.86 & \textbf{0.87} & X & X & X & X & 19.11 & 1.58 & 
		\textbf{8.08} & 0.88\\
		& ORB & \textbf{5.52} & \textbf{0.86} & 74.50 & 38.13 & X & X & 15.07 & 
		2.70 & 5.63 & 1.34\\
		\hline
		\multirow{2}{*}{07}
		& DSO & 6.38 & 2.38 & 6.46 & \textbf{2.29} & 6.34 & 2.38 & 6.21 & 2.34 
		& \textbf{4.55} & 2.36\\
		& ORB & 6.35 & \textbf{2.58} & 94.40 & 9.48 & 67.10 & 30.52 & 65.98 & 
		41.08 & \textbf{4.92} & 2.82\\
		\hline
		\multirow{2}{*}{09}
		& DSO & 7.87 & 4.96 & X & X & X & X & 12.78 & \textbf{2.83} & 
		\textbf{5.57} & 3.16\\
		& ORB & 14.16 & 9.21& 67.10 & 42.44 & X & X & 33.35 & 24.50 & 
		\textbf{5.39} & \textbf{3.78}\\
		\hline
		\hline
		\multirow{2}{*}{mean}
		& DSO & 11.21 & 3.28 & X & X & X & X & 12.35 & 2.44 & \textbf{5.67} & 
		\textbf{2.36}\\
		& ORB & 16.94 & 5.69 & 76.32 & 32.15 & 73.58 & 29.12 & 53.83 & 25.54 & 
		\textbf{6.83} & \textbf{2.92}\\
		\hline
	\end{tabular}
	\caption{Evaluation on the test sequences from the Oxford RobotCar dataset. 
		$t_{rel}(\%)$ and $r_{rel}(^\circ)$ are the relative translational and 
		rotational errors~\cite{geigerwe}, respectively. X means lost tracking 
		and 
		the sequences which lose tracking 
		are not used for calculating the mean. Overall, MFGAN improves 
		both Stereo DSO and Stereo ORB-SLAM in terms of average $t_{rel}$. 
		MFGAN 
		also shows the superior results to other variants of the models.}
	\label{tab:oxford_result}
\end{table}

\begin{figure*}[t]
\begin{subfigure}{0.588\linewidth}
\includegraphics[width=\linewidth]{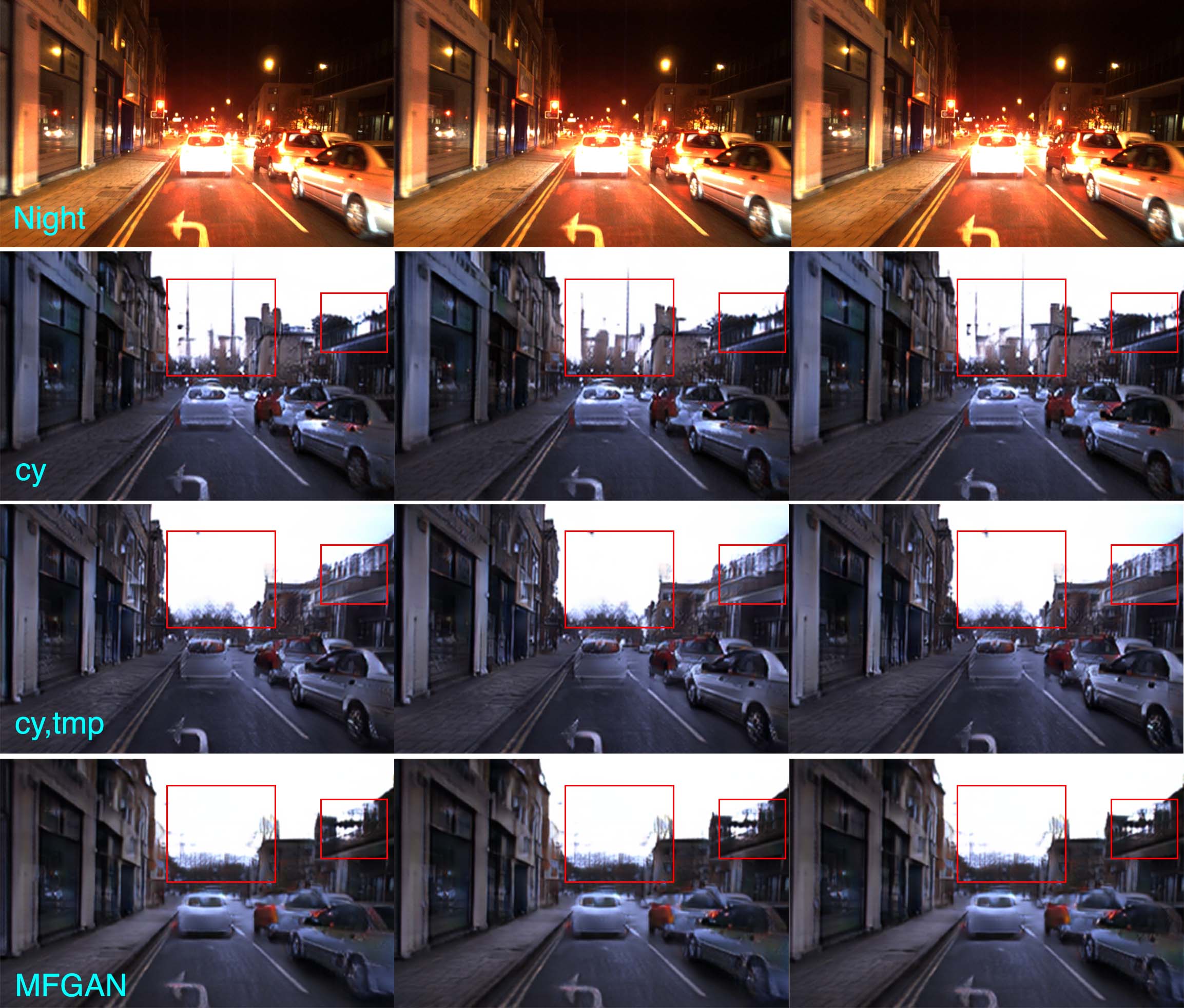}
\caption{}
\label{fig:temp_vis}
\end{subfigure}
\hfill
\begin{subfigure}{0.392\linewidth}
\includegraphics[width=\linewidth]{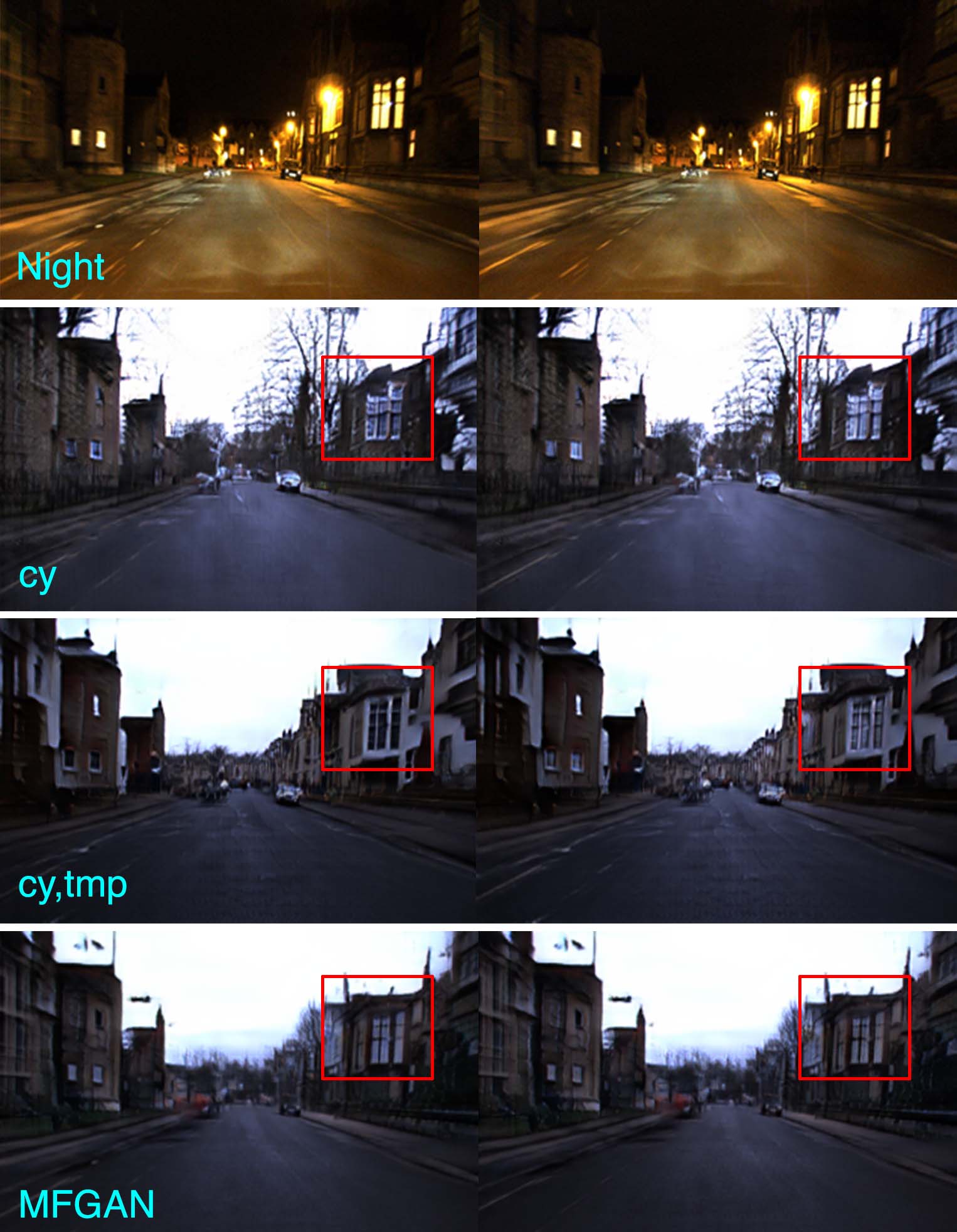}
\caption{}
\label{fig:stereo_vis}
\end{subfigure}
\caption{Qualitative results on temporal consistency (a) and stereo consistency 
(b) for the Oxford RobotCar dataset. Each row shows from top to bottom original 
Night images, the outputs from $cy$, $cy,tmp$, and MFGAN. (a) Left: $t$-th 
frame. Center:
$(t+1)$-th frame. Right: $(t+2)$-th frame. (b) Left: left image of stereo. 
Right: right image of stereo. 
MFGAN presents more coherent appearance regarding temporal as well as 
stereo consistency. Please refer to our supplementary video for clearer 
demonstration.}
\vspace{-6mm}
\label{fig:stereo}
\end{figure*}
%TODO: Upload video.

Therefore, we measure the temporal consistency $E_{tmp}$ by the endpoint error 
~\cite{ilg2017flownet} as below:
\begin{equation}
EPE(W_{t}^{t+2}, W_t^{t+1} \oplus W_{t+1}^{t+2})
\label{eq:epe_temp}
\end{equation}
where $\oplus$ means the addition of two optical flows. $EPE$ indicates the end 
point error which is used to measure the error of two optical flows as 
$\|W_1-W_2\|_2$ where $W_1$ and $W_2$ are flows. The optical flow addition is 
done by adding the first operand to the sampled second operand based on the 
first optical flow.

Likewise, we compute the stereo and temporal consistency $E_{st}$ as below:
\begin{equation}
EPE(W_{t,r}^{t+1,r} \oplus W_{t+1,r}^{t+1,l}, W_{t,r}^{t,l} \oplus 
W_{t,l}^{t+1,l})
\label{eq:epe_stereo}
\end{equation}
where $l,r$ means right and left side of stereo image pair respectively.

%The results of the proposed metric on the New Tsukuba dataset are shown in 
%Table \ref{tab:epe_temporal_tsukuba}.  While the outputs from the $cy$ model 
%show very high EPE in terms of median as well as mean, the outputs from 
%$cy,tmp$ show lower errors than the $cy$ outputs. This presents that adding 
%temporal consistency generates more consistent frames than only using cycle 
%consistency loss.

The results of the temporal and stereo-spatial consistency is shown in 
Table \ref{tab:epe_tmp_st}. While the model $cy,tmp$ gives comparably low EPE 
for temporal consistency $E_{tmp}$, it gives high error values for temporal and 
stereo consistency $E_{st}$. On the other hand, MFGAN, which is trained with 
both temporal and stereo consistency, shows lower EPE than $cy,tmp$. Overall, 
this shows that $cy,tmp$ is more consistent in terms of temporally 
contiguous frames but considering stereo sequences, MFGAN is more consistent 
for the entire stereo sequences.

\textbf{Qualitative results}. The qualitative results of temporal consistency 
converting the Night scene to Day scene of the Oxford RobotCar dataset are 
shown in Figure \ref{fig:temp_vis}. While the generated outputs from the model 
$cy$ give fluctuating artifacts, $cy,tmp$ and MFGAN presents consistent 
appearance over the contiguous frames, especially on the areas marked in red 
rectangle. The comparison for stereo consistency for the same trained model is 
presented in Figure \ref{fig:stereo_vis}. The stereo image pair from MFGAN 
delivers consistent image appearance. Please refer to our 
supplementary video for clearer demonstration.

\subsection{Stereo Visual Odometry}
We run Stereo DSO and stereo ORB-SLAM 5 times for each method and each original 
day sequences (Day), the original night sequences (Night) and the generated Day 
sequences (MFGAN). We use the \textit{median} relative 
translational error $t_{rel}$($\%$), and relative rotational error 
$r_{rel}$($^{\circ}$) as proposed in the KITTI Odometry 
benchmark~\cite{geigerwe} as the 
evaluation metric. Please refer to our 
\href{https://vision.in.tum.de/_media/spezial/bib/jung2019corl-supp.pdf}{supplemental
	material} for the 
formulas of calculating $t_{rel}$ and $r_{rel}$. We run Stereo DSO with the 
default 
settings on all sequences 
and Stereo ORB-SLAM with lower FAST corner thresholds on Night due to low 
gradient magnitude of dark scenes. Both methods 
show in general good performance on Day but declined performance on Night.
With the translated images from MFGAN, both VO 
methods are significantly improved. Please refer to our 
\href{https://vision.in.tum.de/_media/spezial/bib/jung2019corl-supp.pdf}{supplemental
	material} for the evaluation results on Day.

\begin{table*}[t]
	\centering
	\scriptsize
	\begin{tabular}{cc|cc|cc|cc||cc}
		\hline
		& & \multicolumn{2}{c|}{AHE~\cite{zuiderveld1994contrast}} & 
		\multicolumn{2}{c|}{LIME~\cite{guo2016lime}} & 
		\multicolumn{2}{c||}{DP~\cite{deephotoenhancer}} & 
		\multicolumn{2}{c}{MFGAN}\\ 
		Seq. & & $t_{rel}$ & $r_{rel}$ & $t_{rel}$ & $r_{rel}$ & $t_{rel}$ & 
		$r_{rel}$ & 
		$t_{rel}$ & $r_{rel}$ \\
		\hline
		\multirow{2}{*}{01}  
		& DSO & 7.82 & 1.84 & 7.80 & 1.87 & 7.66 & \textbf{1.79} & 
		\textbf{5.41} & 2.18\\
		& ORB & 46.85 & 10.43 & 38.07 & 3.96 & 53.69 & 12.97 & \textbf{11.06} & 
		\textbf{2.75} \\
		\hline
		\multirow{2}{*}{04}
		& DSO & 7.22 & 4.40  & 6.59 & \textbf{2.75} & 6.49 & 2.92 & 
		\textbf{4.75} & 3.20\\
		& ORB & 35.00 & 12.40 & 27.43 & 10.31 & 40.77 & 17.41 & \textbf{4.59} & 
		\textbf{3.89}\\
		\hline
		\multirow{2}{*}{06}
		& DSO & 10.08 & \textbf{0.73} & 9.77 & 0.80 & 9.63 & 0.79 & 
		\textbf{8.08} & 0.88\\
		& ORB & 6.56 & 1.40 & \textbf{5.69} & \textbf{0.89} & 34.87 & 0.93 & 
		5.67 & 1.34\\
		\hline
		\multirow{2}{*}{07}
		& DSO & 6.46 & \textbf{2.29} & 6.34 & 2.38 & 6.21 & 2.34 & 
		\textbf{4.55} & 2.36\\
		& ORB & 11.91 & 3.06 & 5.50 & 2.57 & 24.48 & \textbf{2.28} & 
		\textbf{4.92} & 2.82\\
		\hline
		\multirow{2}{*}{09}
		& DSO & 6.85 & 3.22 & 6.51 & 2.82 & 6.49 & \textbf{2.73} & 
		\textbf{5.57} & 3.16\\
		& ORB & 24.97 & 13.81 & 17.32 & 13.20 & 24.39 & 4.11 & \textbf{5.39} & 
		\textbf{3.78}\\
		\hline
		\hline
		\multirow{2}{*}{mean}
		& DSO & 7.69 & 2.50 & 7.40 & \textbf{2.12} & 7.30 & \textbf{2.12} & 
		\textbf{5.67} & 2.36\\
		& ORB & 25.06 & 8.22 & 18.80 & 6.19 & 35.64 & 7.54 & \textbf{6.83} & 
		\textbf{2.92}\\
		\hline
	\end{tabular}
	\caption{Comparison with other photo enhancing methods.}
	\label{tab:oxford_benchmark_photo_enhance}
\end{table*}

\begin{figure}[t]
	\begin{subfigure}{0.5\linewidth}
		\includegraphics[width=\linewidth]{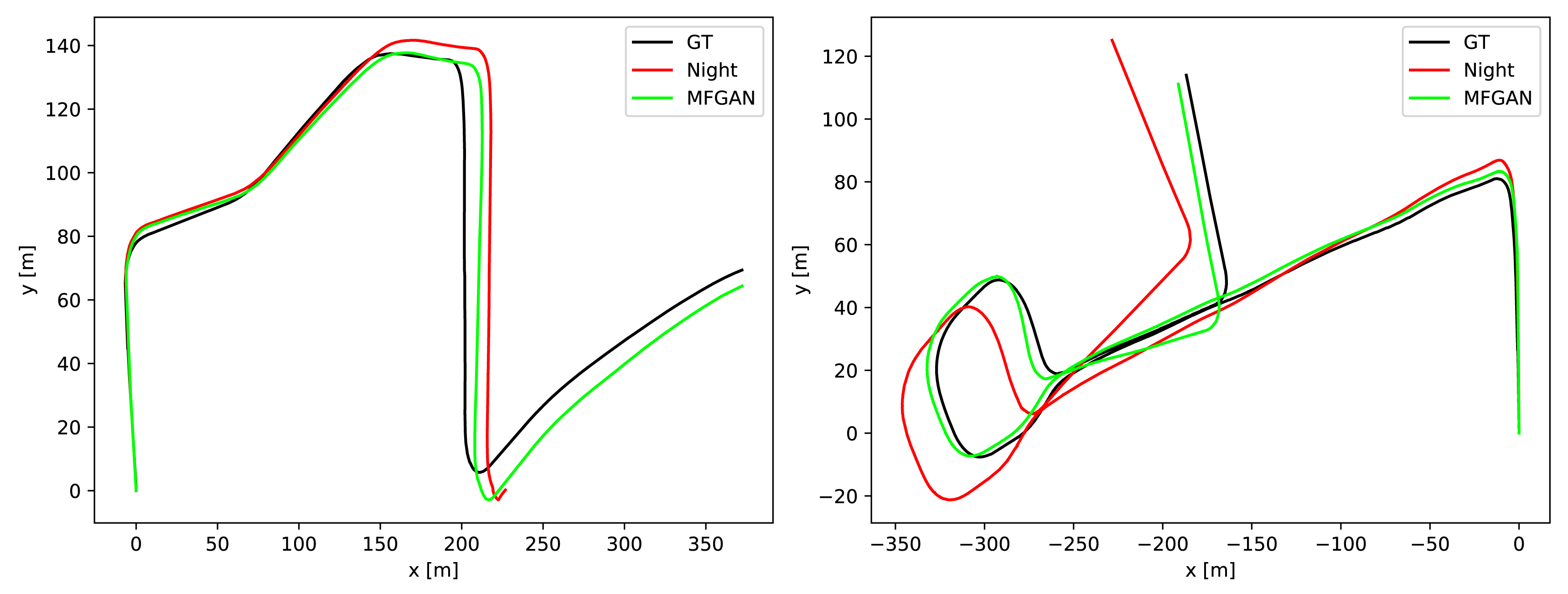}
		\caption{Stereo DSO}
		\label{fig:dso_traj}
	\end{subfigure}
	\hfill
	\begin{subfigure}{0.5\linewidth}
		\includegraphics[width=\linewidth]{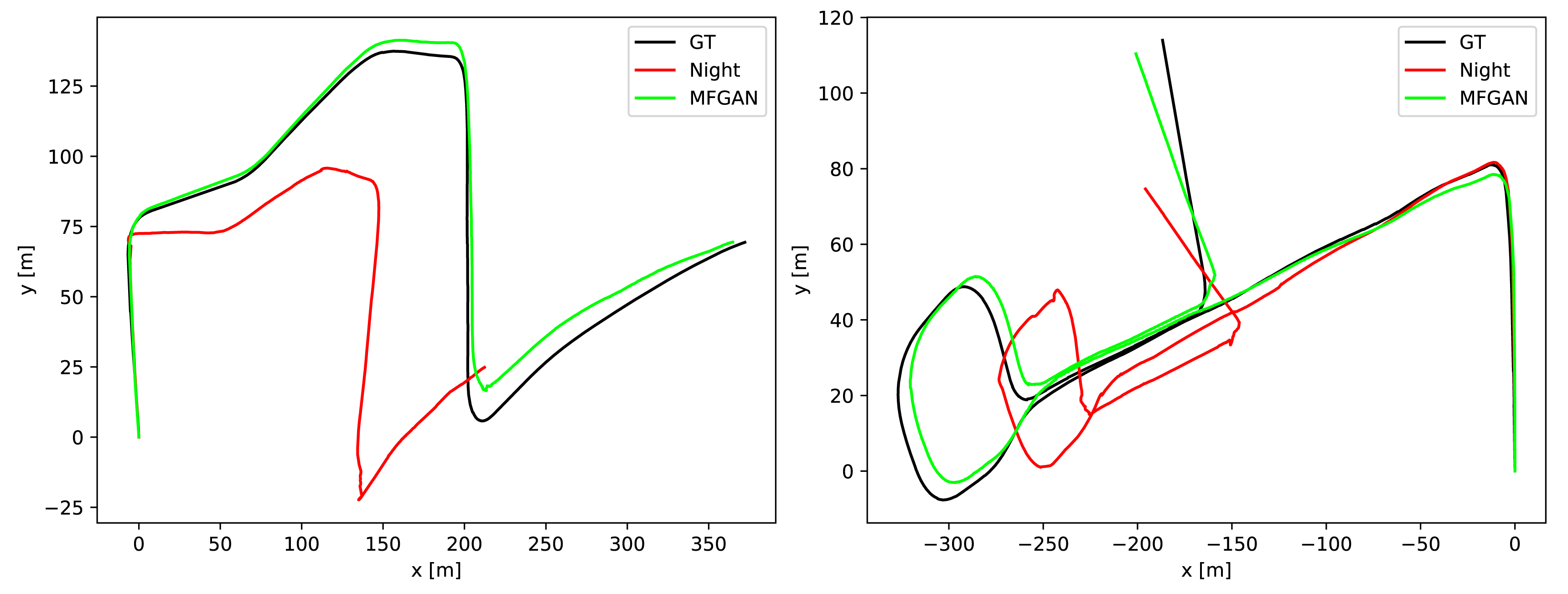}
		\caption{Stereo ORB-SLAM}
		\label{fig:orb_traj}
	\end{subfigure}
	\caption{Results of Stereo VO methods on Seq. 4 and 9 of the Oxford 
		RobotCar 
		dataset. Clearly MFGAN improves the 
		performance of both 
		Stereo DSO and 
		Stereo ORB-SLAM. Please refer to our 
		\href{https://vision.in.tum.de/_media/spezial/bib/jung2019corl-supp.pdf}{supplemental
			material} the 
		trajectories 
		of other sequences.}
	\label{fig:traj}
	\vspace{-7mm}
\end{figure}

%\begin{figure}
%	\includegraphics[width=0.5\linewidth,]{./figures/traj.png}
%	\caption{Results of Stereo DSO  and stereo ORB-SLAM on Oxford RobotCar 
%		dataset. From the top to the last row, each trajectory is Seq. 2, 4, 9. 
%		Clearly MFGAN improves the performance of both indirect and direct 
%		methods.}
%	\label{fig:traj}
%\end{figure}	

%The evaluation of MFGAN on Oxford RobotCar dataset is shown in 
%Table~\ref{tab:oxford_result}. We use the relative error metric 
%proposed in\todo{citeKITTI} and show the results on the testing sequences. 
%Our method significantly improves the tracking performance for 
%the sub-sequences which show worse performance at night than at day time. With 
%both direct and indirect method, the model $cy$ gives unstable results as the 
%VO methods losing tracking for many sequences. The model $cy,tmp$ shows 
%slightly better performance than the model $cy$, but still especially stereo 
%ORB-SLAM loses tracking in several sequences. This result reveals that in 
%addition to the advantage of temporal consistency over contiguous frames, 
%stereo consistency between stereo image pair is beneficial in terms of stereo 
%matching in stereo VO algorithms, and allow us to fully exploit the 
%functionality of stereo VO methods with the enhanced image sequences.

\begin{figure}[t]
	\centering
	\includegraphics[width=.3\linewidth]{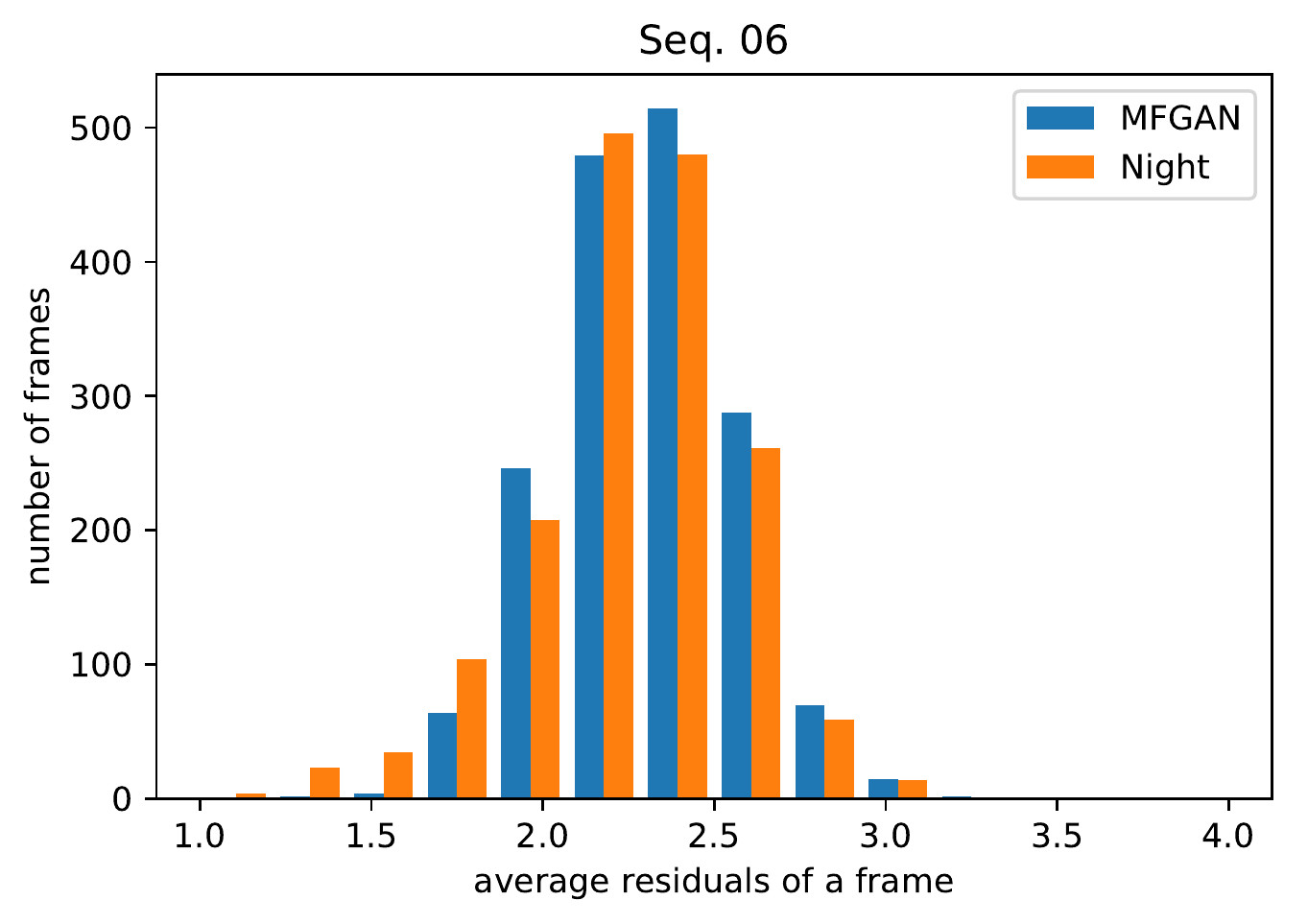}
	\includegraphics[width=.3\linewidth]{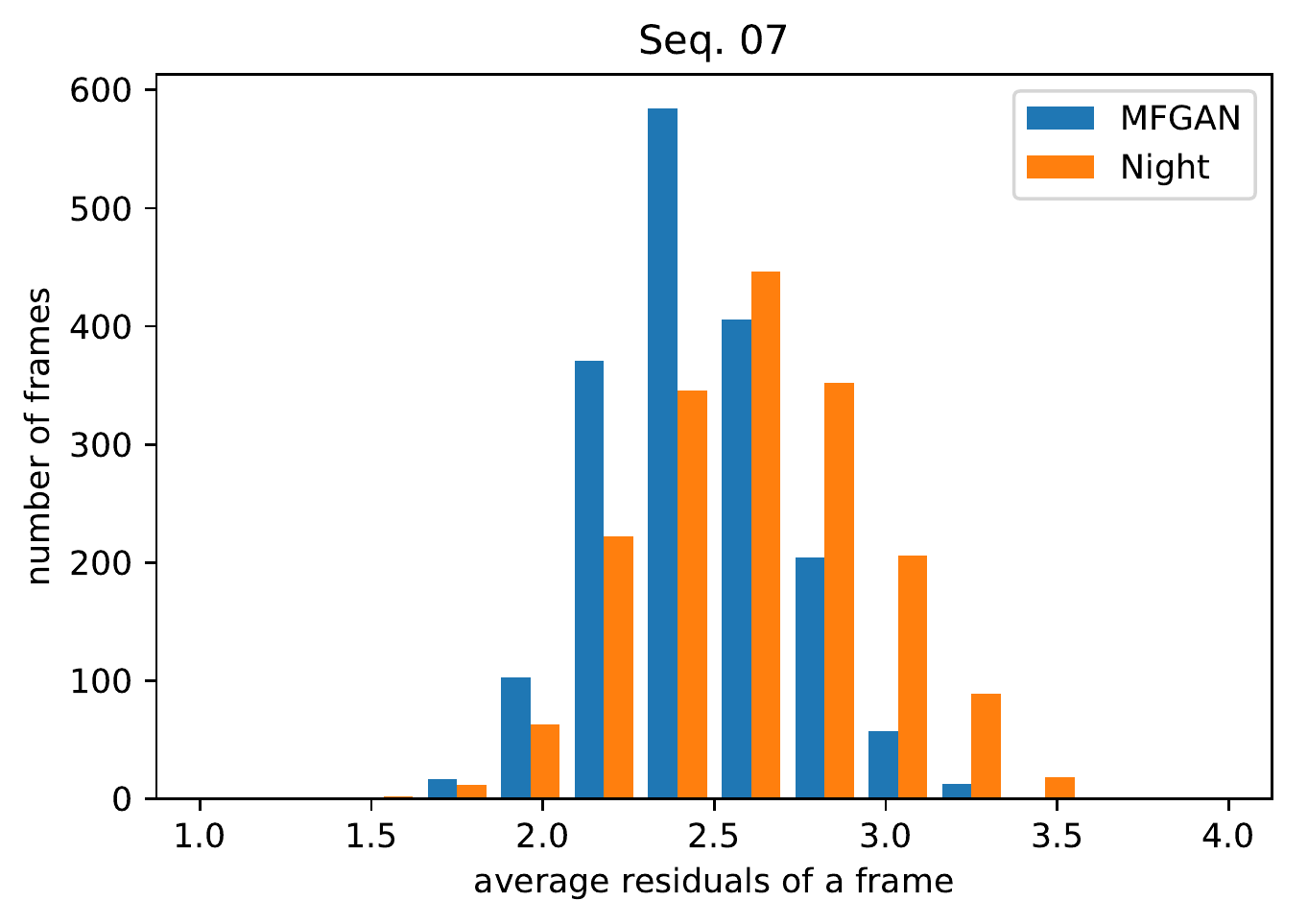}
	\includegraphics[width=.3\linewidth]{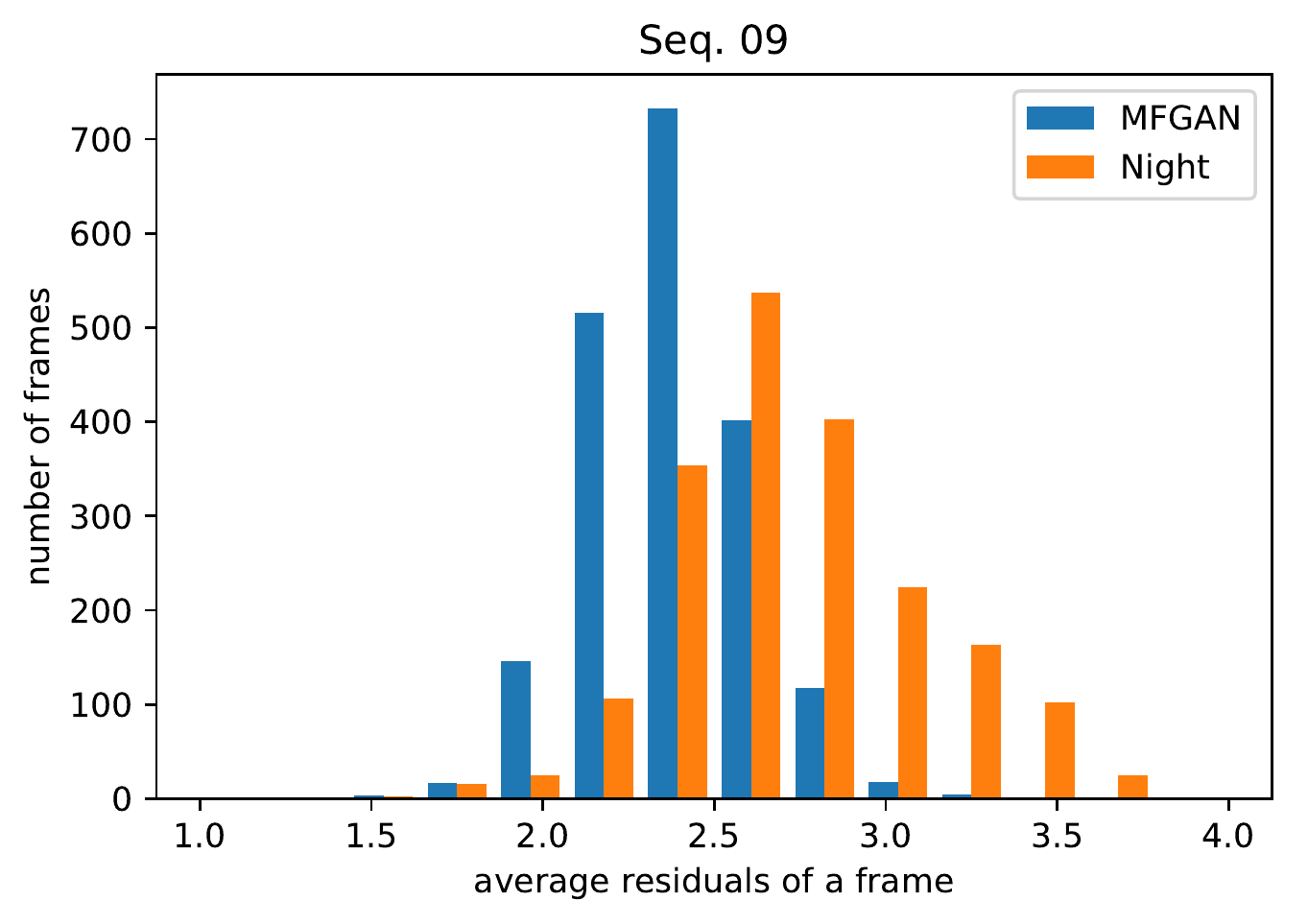}
	\caption{Histogram of the average frame residuals.}
	\label{fig:hist}
\end{figure}

%\subsubsection{Comparison with original night sequences}
As shown in Table~\ref{tab:oxford_result}, both Stereo DSO and 
Stereo ORB-SLAM are improved on MFGAN on all the sequences except for 
Seq. 06 for which Stereo ORB-SLAM can already deliver very accurate tracking on 
Night. The results from 
$cy$, ${cy,tmp}$, ${cy,st}$ and MFGAN show the effectiveness 
of the proposed consistency loss terms and only with our full approach, MFGAN, 
the performance is improved consistently.
Overall Stereo DSO delivers 
better results than Stereo ORB-SLAM on Night, and the 
improvement from MFGAN is more significant for Stereo ORB-SLAM.
%The reason is that Stereo 
%DSO, as a direct method, is able to track more points than Stereo 
%ORB-SLAM, whereas ORB-SLAM cannot track enough points, despite using lower 
%FAST 
%thresholds.
The results of Seq. 01 and 04 on which Stereo ORB-SLAM loses tracking on Night 
show that MFGAN improves the robustness of Stereo ORB-SLAM. With the 
day-ification from MFGAN, the \textit{quantity} of tracked points for each 
frame increase from $106$ to $289$, which improves the robustness of ORB-SLAM.
%Surprisingly, on \todo{some sequences}, the 
%performance of the VO methods is comparable on both the day and night scene. 
%Our observation is that the corresponding sequences contain fairly good 
%lighting conditions, e.g., enough streetlights, less active lighting /
%halo, \textit{and} even less moving objects.
%\textbf{Further discussion}.
The tracking accuracy of Stereo DSO 
is improved on MFGAN, since the brightness is more consistent by 
removing the active lighting / halo while preserving the coherence of 
consecutive frames, which reduces the amount of outliers for the photometric 
error minimization. For ORB-SLAM, on Seq. 07 and 09, the 
tracking of Stereo ORB-SLAM does not fail on Night and the amounts of 
tracked points are similar for Night and MFGAN -- in 
average, 289 and 293 points are tracked for each frame, respectively. 
Therefore, we also 
measure the \textit{quality} of the features by taking the average 
re-projection residuals of the local map points for each frame when the pose 
bundle adjustment is finished. Please refer to~\cite{mur2017orb} 
for the details of the optimization. The histograms of the residuals for 
Seq. 07 and 09 are shown in Figure~\ref{fig:hist}. We can see that 
on MFGAN Stereo ORB-SLAM can deliver more points with lower 
re-projection residuals and less points with higher residuals, which improves 
the tracking accuracy. The higher average residuals from Night 
indicates that there are more wrong matches due to the active lighting or 
inaccurate matches caused by less reliable descriptors due to the dark scene 
with less image gradients. As a comparison, we also show the histogram of Seq. 
06 in Figure~\ref{fig:hist}.

In Table~\ref{tab:oxford_benchmark_photo_enhance}, we compare MFGAN with other 
photo enhancing methods including adaptive histogram 
equalization(AHE)~\cite{zuiderveld1994contrast}, low-light image 
enhancement(LIME)~\cite{guo2016lime}, deep photo 
enhancer(DP)~\cite{deephotoenhancer}. We use the 
Matlab implementation for AHE 
and LIME, and the pre-trained model for DP. From the table we can see that 
MFGAN is able to deliver consistent improvement for both Stereo DSO and Stereo 
ORB-SLAM on all the sequences.

To show the advantage of MFGAN for VO compared with other recent style transfer 
methods, in Table~\ref{tab:oxford_benchmark_photo_enhance}, we show the results 
obtained by using the translated sequences from 
ToDayGAN~\cite{anoosheh2018night}, DRIT~\cite{DRIT}, and 
LinearTransfer(LT)~\cite{li2018learning}. For ToDayGAN we use the 
pre-trained model, since it also trained for Day-Night style transfer. We train 
DRIT with our split and do the inference with a constant noise vector. For LT, 
we use a Day image as the style, and run the video inference on the test 
sequences. From the table we can see that MFGAN outperforms other methods in 
terms of both accuracy and robustness for VO.

\begin{table*}[t]
	\centering
	\scriptsize
	\begin{tabular}{cc|cc|cc|cc||cc}
		\hline
		& & \multicolumn{2}{c|}{ToDayGAN~\cite{anoosheh2018night}} & 
		\multicolumn{2}{c|}{DRIT~\cite{DRIT}} & 
		\multicolumn{2}{c||}{LT~\cite{li2018learning}} & 
		\multicolumn{2}{c}{MFGAN}\\ 
		Seq. & & $t_{rel}$ & $r_{rel}$ & $t_{rel}$ & $r_{rel}$ & $t_{rel}$ & 
		$r_{rel}$ & 
		$t_{rel}$ & $r_{rel}$ \\
		\hline
		\multirow{2}{*}{01}  
		& DSO & 77.6 & 9.76 & 82.57 & 10.18 & 7.22 & \textbf{2.08} & 
		\textbf{5.41} & 2.18\\
		& ORB & 82.97 & 26.22 & X & X & 11.78 & 2.92 & \textbf{11.06} & 
		\textbf{2.75} \\
		\hline
		\multirow{2}{*}{04}
		& DSO & 74.69 & 15.19 & 74.71 & 11.40 & 6.05 & 3.23 & \textbf{4.75} & 
		\textbf{3.20}\\
		& ORB & 75.13 & 38.19 & X & X & 12.62 & 7.41 & \textbf{4.59} & 
		\textbf{3.89}\\
		\hline
		\multirow{2}{*}{06}
		& DSO & 13.90 & 2.61 & 7.73 & 1.18 & 9.46 & 0.97 & \textbf{8.08} & 
		\textbf{0.88}\\
		& ORB & 97.76 & 9.55 & 98.69 & 10.07 & 5.85 & \textbf{0.80} & 
		\textbf{5.67} & 1.34\\
		\hline
		\multirow{2}{*}{07}
		& DSO & 21.47 & 3.55 & 67.55 & 5.99 & X & X & \textbf{4.55} & 
		\textbf{2.36}\\
		& ORB & 67.80 & 42.44 & 68.30 & 42.97 & X & X & \textbf{4.92} & 
		\textbf{2.82}\\
		\hline
		\multirow{2}{*}{09}
		& DSO & 36.20 & 10.60 & 29.79 & 12.83 & 6.23 & 3.20 & \textbf{5.57} & 
		\textbf{3.16}\\
		& ORB & 62.84 & 44.51 & 62.60 & 45.61 & 10.77 & 7.78 & \textbf{5.39} & 
		\textbf{3.78}\\
		\hline
		\hline
		\multirow{2}{*}{mean}
		& DSO & 44.77 & 8.34 & 52.47 & 8.32 & 7.85 & \textbf{2.08}& 
		\textbf{5.67} & 2.36\\
		& ORB & 77.30 & 32.18 & 76.53 & 32.88 & 8.31 & 4.29 & \textbf{6.83} & 
		\textbf{2.92}\\
		\hline
	\end{tabular}
	\caption{Comparison with other style transfer methods.}
	\label{tab:oxford_benchmark_style_transfer}
\end{table*}

% !TeX root = ../main.tex

\section{Conclusion}

In this paper, we present a learning-based approach to improve stereo VO 
methods in low lighting conditions. To this end, we introduced the concept of  
Multi-Frame GAN (MFGAN) which performs a spatio-temporally consistent domain 
transfer. MFGAN takes advantage of 
unpaired datasets by leveraging a novel cycle adversarial network and learns to 
generate frames with temporal and stereo coherence. With the proposed metric 
for frame consistency, we 
quantitatively validate that our method successfully generates images with 
temporal as well as stereo consistency. Experiments regarding VO on 
both a synthetic indoor dataset and a real outdoor dataset show that 
our method improves the performance of both indirect and direct VO methods 
in low light environments. We also show that MFGAN outperforms other photo 
enhancement and 
image/video translation methods by a notable margin. In future work, we will 
explore the generalization 
capability of the proposed temporal and stereo consistency loss on other style 
transfer methods.
% !TeX root = ../main.tex

\section{Acknowledgments}

We would like to thank the reviewers for their constructive comments which help 
us improve the paper. We also would like to thank the researchers from 
Artisense for the proofreading as well as the fruitful discussions.

%===============================================================================

%\newpage
\bibliography{example}  % .bib

\end{document}